\documentclass{article}

    \PassOptionsToPackage{numbers}{natbib}

\usepackage[final]{neurips_2020_ml4ps}
\usepackage{xcolor}
\usepackage{amsmath}
\usepackage[utf8]{inputenc} 
\usepackage[T1]{fontenc}    
\usepackage[hidelinks]{hyperref}       
\usepackage{url}            
\usepackage{booktabs}       
\usepackage{nicefrac}       
\usepackage{microtype}      
\usepackage{bm}
\usepackage{graphicx,xcolor}
\usepackage[framemethod=tikz]{mdframed}
\usepackage{caption}
\usepackage{subcaption}
\usepackage{enumitem}
\usepackage[textsize=small]{todonotes}
\usepackage[linesnumbered,ruled]{algorithm2e}
\title{Spacecraft Collision Risk Assessment with\\ Probabilistic Programming}

\author{%
\begin{minipage}[t]{0.466\textwidth}
  \begin{center}
  Giacomo Acciarini\\
  \normalfont{University of Strathclyde}\\
  \texttt{giacomo.acciarini@gmail.com}
  \end{center}
\end{minipage}
  \And
\begin{minipage}[t]{0.466\textwidth}
  \begin{center}
  Francesco Pinto\\
  \normalfont{University of Oxford}\\
  \texttt{francesco.pinto@eng.ox.ac.uk}
  \end{center}
\end{minipage}
  \AND 
\begin{minipage}[t]{0.466\textwidth}
  \begin{center}
  Sascha Metz\\
  \normalfont{Technische Universität Darmstadt}\\
  \texttt{smetz1405@gmail.com}
  \end{center}
\end{minipage}
  \And
\begin{minipage}[t]{0.466\textwidth}
  \begin{center}
  Sarah Boufelja\\
  \normalfont{IBM}\\
  \texttt{boufelja.sarah@gmail.com}
  \end{center}
\end{minipage}
  \AND
\begin{minipage}[t]{0.466\textwidth}
  \begin{center}
  Sylvester Kaczmarek\\
  \normalfont{Imperial College London}\\
  \texttt{sylvester.kaczmarek@gmail.com}
  \end{center}
\end{minipage}
  \And
\begin{minipage}[t]{0.466\textwidth}
  \begin{center}
  Klaus Merz\\
  \normalfont{European Space Agency}\\
  \texttt{klaus.merz@esa.int}
  \end{center}
\end{minipage}
  \AND
\begin{minipage}[t]{0.466\textwidth}
  \begin{center}
  José A.~Martinez-Heras\\
  \normalfont{European Space Agency}\\
  \texttt{jose.antonio.martinez.heras@esa.int}
  \end{center}
\end{minipage}
  \And   
\begin{minipage}[t]{0.466\textwidth}
  \begin{center}
  Francesca Letizia\\
  \normalfont{European Space Agency}\\
  \texttt{francesca.letizia@esa.int}
  \end{center}
\end{minipage}
  \AND
\begin{minipage}[t]{0.466\textwidth}
  \begin{center}
  Christopher Bridges\\
  \normalfont{University of Surrey}\\
  \texttt{c.p.bridges@surrey.ac.uk}
  \end{center}
\end{minipage}
  \And 
\begin{minipage}[t]{0.466\textwidth}
  \begin{center}
  Atılım Güneş Baydin\\
  \normalfont{University of Oxford}\\
  \texttt{gunes@robots.ox.ac.uk}
  \end{center}
\end{minipage}

}

\begin{document}

\maketitle

\begin{abstract}

Over 34,000 objects bigger than 10 cm in length are known to orbit Earth. Among them, only a small percentage are active satellites, while the rest of the population is made of dead satellites, rocket bodies, and debris that pose a collision threat to operational spacecraft. Furthermore, the predicted growth of the space sector and the planned launch of megaconstellations will add even more complexity, therefore causing the collision risk and the burden on space operators to increase. Managing this complex framework with internationally agreed methods is pivotal and urgent. In this context, we build a novel physics-based probabilistic generative model for synthetically generating conjunction data messages, calibrated using real data. By conditioning on observations, we use the model to obtain posterior distributions via Bayesian inference. We show that the probabilistic programming approach to conjunction assessment can help in making predictions and in finding the parameters that explain the observed data in conjunction data messages, thus shedding more light on key variables and orbital characteristics that more likely lead to conjunction events. Moreover, our technique enables the generation of physically accurate synthetic datasets of collisions, answering a fundamental need of the space and machine learning communities working in this area.

\end{abstract}

\section{Introduction}

The low Earth orbit (LEO) environment is becoming increasingly crowded \cite{lemmens2020esaannual}, with an estimate of more than half a million uncontrolled orbiting objects bigger than 1 cm in length.\footnote{\url{https://sdup.esoc.esa.int/discosweb/statistics/} (August 2020)} Furthermore, the space population is expected to increase more in the future, due to the expansion of the space sector and the advent of megaconstellations \cite{kennewell2013overview,muelhaupt2019space,radtke2017interactions,rossi2017quantitative}. This scenario poses a danger to operational satellites, scientific missions, and the future of human access to space since the risk of collision among spacecraft will increase, with an increased risk of a cascade of collisions that could affect the whole space population---the Kessler syndrome \cite{kessler1978collision}. Moreover, the burden on human operators that have to maneuver these satellites can become complex and unmanageable \cite{merz2017current}. Several guidelines have been developed by the Inter-Agency Space Debris Coordination Committee (IADC) in recent years to mitigate collision risk in the future \cite{inter2007iadc}. Meanwhile, space agencies, private operators and the research community have been investigating ways to tackle this problem \cite{walker2004cost,krag2012consideration,vasile2018artificial,zuiani2012preliminary,kumar2015agora,visagie2015drag}. Current collision avoidance strategies are based on techniques that assess the risk of collision between pairs of objects, starting from a series of conjunction data messages (CDMs) associated with a potential collision event. 

A CDM is automatically issued by the US-based Combined Space Operations Center (CSpOC), which uses the data provided by the US Strategic Command (USSTRATCOM) that routinely observes and tracks objects in space via a global Space Surveillance Network (SSN). The CDMs are issued to the owner/operator (O/O) of the involved spacecraft. Usually, they contain the following information about a potential collision event: (1) the predicted orbital position and velocity of the two objects at their time of closest approach (TCA); (2) information related to the physical simulation environment used for making the predictions; (3) the propagated uncertainty information about the two objects' physical states, in the form of covariances. Additionally, they might also optionally contain other details about the probability of collision and the method used to compute it. The data contained in CDMs are calculated by CSpOC through a range of different algorithms, involving the propagation of uncertainties from current observational data up until the time of closest approach \cite{vallado2001fundamentals,sun2019nonlinear,jones2013nonlinear,vasile2019set,adurthi2015conjugate}.  
Once uncertainties are propagated to TCA, several algorithms can be used for the evaluation of collision risk (i.e., the probability that the two objects will collide at TCA) \cite{alfriend1999probability,patera2001general, klinkrad2010space,patera2003satellite,chan2008spacecraft}. The operators then use this information combined with their own risk assessment and data (e.g., telemetry) to decide whether to undertake a maneuver to avoid the conjunction.

Usually, for each collision event a time series of CDMs covering one week is released. At the European Space Operations Center (ESOC) of the European Space Agency (ESA), CDM updates are received roughly every eight hours. However, they do not necessarily contain updated information for both the objects, since very often only one of the two objects (the \emph{target}, often an operational satellite) is well observed, whereas the other one (the \emph{chaser}, often a piece of debris) is not \cite{merz2017current}. Target and chaser objects have different frequency of observation updates and uncertainty characteristics, creating a complex setting where the tools of probabilistic machine learning (ML) and simulation-based inference \cite{cranmer2020frontier} are applicable. A major obstacle in applying ML in this setting is the scarcity of publicly available data, where space agencies are reluctant to share data due to security concerns, and a severe class imbalance in any conjunction dataset as collisions are rare events.

In this work, we build a physics-based probabilistic generative model to simulate spacecraft orbits, observations, and the CDM generation process, using probabilistic programming \cite{IntroProbProg}. We calibrate the model using real data, and use it to perform Bayesian inference. We show that this approach could play a pivotal role in helping operators determine the key variables that lead to conjunction events and to make predictions that are reliable (i.e., with associated uncertainties) and explainable (in the context of model-based Bayesian inference). The approach also enables the sampling of large datasets of synthetic but realistic conjunction events that can deliver much-needed publicly available data for the ML and space communities to develop and evaluate new techniques.

\section{Probabilistic Programming}
Probabilistic programming is a paradigm that allows to specify probabilistic models for data generation and to perform inference in the model conditioned on observed data \cite{wood2014new,IntroProbProg}. Using a universal programming language enhanced by two constructs (\texttt{sample} and \texttt{observe}), a domain expert can describe a system's behavior using general-purpose (Turing-complete) programming languages, essentially creating a stochastic simulator in which one can perform automated Bayesian inference \cite{etalumis}. We denote the observable data we want to perform inference on as $\mathbf{y}$, and assume that $\mathbf{y}$ can be generated by executing a generative program that depends on the values sampled from some latent variables $\mathbf{x}$. A probabilistic program specifies the joint probability distribution $p(\mathbf{x},\mathbf{y}) = p(\mathbf{y}\vert\mathbf{x})\,p(\mathbf{x}) $ in terms of the prior distributions $p(\mathbf{x})$ over the latent variables $\mathbf{x}$ (using the \texttt{sample} statements), an interpretable procedure that computes $\mathbf{y}$ given $\mathbf{x}$, and a likelihood distribution $p(\mathbf{y}\vert\mathbf{x})$ (using the \texttt{observe} statements). The model can be run forward by sampling $\mathbf{x}$ from the priors and applying the generative procedure to generate some data $\mathbf{y}$. However, probabilistic programming also allows to invert the generative process to perform inference, i.e., given some observed data $\mathbf{y}$, infer distributions over the latent variables $\mathbf{x}$ that could have generated $\mathbf{y}$. In other words, inference aims to compute the posterior distribution $p(\mathbf{x}\vert\mathbf{y})$, which represents our updated belief about the latents $\mathbf{x}$ after incorporating the information from observed data $\mathbf{y}$.

The order in which the execution of a probabilistic program encounters \texttt{sample} statements may be different from one execution to another. Hence, a way to uniquely identify each of their executions is to represent the execution trace of a probabilistic program as a sequence $\mathcal{T}= \{(x_t,a_t,i_t)\}^T_{t=1}$, where $x_t$, $a_t$ and $i_t$ respectively represent the sampled value, the lexical address of the statement in the program and the instance (i.e. a counter recording the number of times the \texttt{sample} statement at address $a_t$ has been called until the time step $t$) of the $t-$th entry in the trace. The trace length $T$ and ordering can change at each execution. For each trace, we can collect the $T$ sampled values in a sequence $\mathbf{x} := \{x_t\}^T_{t=1}$. Assuming, for notation simplicity, that the order in which the \texttt{observe} statements are encountered is fixed, we can denote the observed values as $\mathbf{y} := \{(y_n)\}_{n=1}^N$. 
 
The joint probability density of an execution trace is therefore $p(\mathbf{x},\mathbf{y}) := \prod_{t=1}^T f_{a_t}(x_t\vert x_{1:t-1}) \prod_{n=1}^N g_n(y_n\vert x_{1:\tau(n)}),$ where $f_{a_t}$ is the probability distribution specified by the \texttt{sample} statement $d_t$ at address $a_t$ (i.e., the prior conditional density given the sample values $x_{1:t-1}$ before encountering the $t-$th \texttt{sample} statement) and $g_n$ is the probability distribution specified by the $n-$th \texttt{observe} statement (i.e., the likelihood density given the sample values $x_{1:\tau(n)}$ obtained before encountering the $n-$th \texttt{observe} statement, where $\tau$ maps the index $n$ of the \texttt{observe} statement to the index of the last \texttt{sample} statement encountered before the $n-$th \texttt{observe} statement during the execution of the program).
In this framework, performing inference means computing an approximation of the posterior $p(\mathbf{x}\vert\mathbf{y})$ and the expectations over chosen functions $\xi$ under the posterior, i.e., $I_\xi = \int \xi(\mathbf{x})p(\mathbf{x}\vert\mathbf{y})d\mathbf{x}$. Many inference procedures are available for this purpose \cite{Wingate2011, Ritchie2016b, Wood2014,Rainforth2016, IC, ICAttention}. For the preliminary results presented in this paper we use importance sampling (likelihood weighting) \cite{IntroProbProg}.

\section{Probabilistic Model of Orbits, Conjunctions, and CDM Generation}
 
\begin{figure}
\centering
\includegraphics[width=1\textwidth]{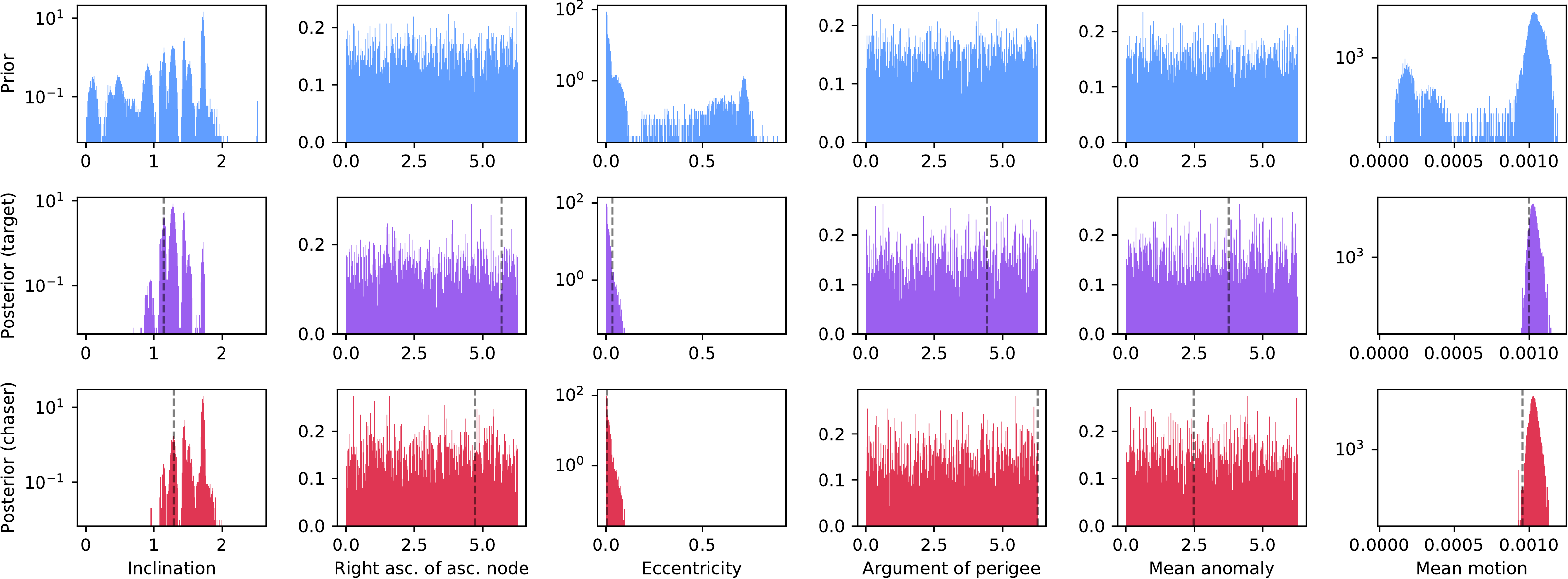}
\caption{\emph{Top:} a part of the prior distributions for orbital elements of LEO population derived from real object catalogs. \emph{Middle and bottom:} posterior distributions of a subset of latent variables for target and chaser. The posterior is conditioned on a ground truth test event, here highlighted with a dashed vertical line, which was generated through a forward run of the simulator.}
\label{fig:dists}
\end{figure}

The proposed stochastic generative model includes the orbital mechanics of the target and chaser and a CDM generation pipeline (incorporating domain knowledge) that mimics the real-world CDM issuing process. An essential part of our probabilistic model are the distributions characterizing the objects and their orbits. We first generate ground truth trajectories of chaser and target by sampling the initial conditions (i.e., initial \emph{two-line elements} \cite{tle}) from prior distributions of the LEO population that we constructed based on real data from the public USSTRATCOM catalog and ESA DISCOS database\footnote{\url{https://discosweb.esoc.esa.int/}} (Figure~\ref{fig:dists}), and instantiating the objects. We then simulate the objects for seven days using physics-based orbit propagators, which model the object's state (position and velocity) taking factors such as drag, radiation, and gravitation effects from planetary bodies into account.

Once this is achieved, a check is made to inspect if a conjunction alert is to be issued (by checking if the distance between the two objects is less than a threshold of 5 km). If that is true then the event is flagged as a conjunction and further monitoring is required leading to a series of generated CDMs. In this work, we calibrated the satellite measurement errors using anonymized real conjunction data released by ESA as the Kelvins challenge.\footnote{\url{https://kelvins.esa.int/collision-avoidance-challenge/data/} (September 2020)} For doing this, we tuned the initial observation measurement noise up until the propagated covariance values at TCA were matching the data from Kelvins. This calibration was done by inspection (i.e., checking whether the distributions of covariances overlapped) and we plan to assess this in a more systematic fashion in future work. Besides the covariances, one could think of calibrating the probabilistic generative model with other real CDM data. However, the main hindrance for a full calibration of the model is caused by the lack of publicly available data, which can only be addressed by collaborating with an owner/operator of satellites.

For issuing CDMs, the state of target and chaser are observed with some associated uncertainty (that vary depending on the sensors used to observe the objects, such as GPS, radar, etc.), then a series of orbit simulation are performed to propagate the state and the associated uncertainties to a future time range. Not only the nominal state, but also the associated initial uncertainties (due to observation errors) are propagated at future times by using a sample-based Monte Carlo approach \cite{yanez2019gaussianity}.

Therefore, the essential elements for a probabilistic generative model of CDMs are: the initial state of both chaser and target, the dynamical environment characteristics (that depend on the simulator used for orbit propagation), the associated observation uncertainties (that are propagated into the future) and the method used for propagating them until time of closest approach. For our preliminary experiments we use SGP4 for orbit propagation, a fast low-precision propagator \cite{hoots1980models}: this is a well known propagator often used for preliminary orbit analysis in LEO, when two-line elements data is available. 

\section{Experiments}

\begin{figure}
\centering
\includegraphics[width=0.99\textwidth]{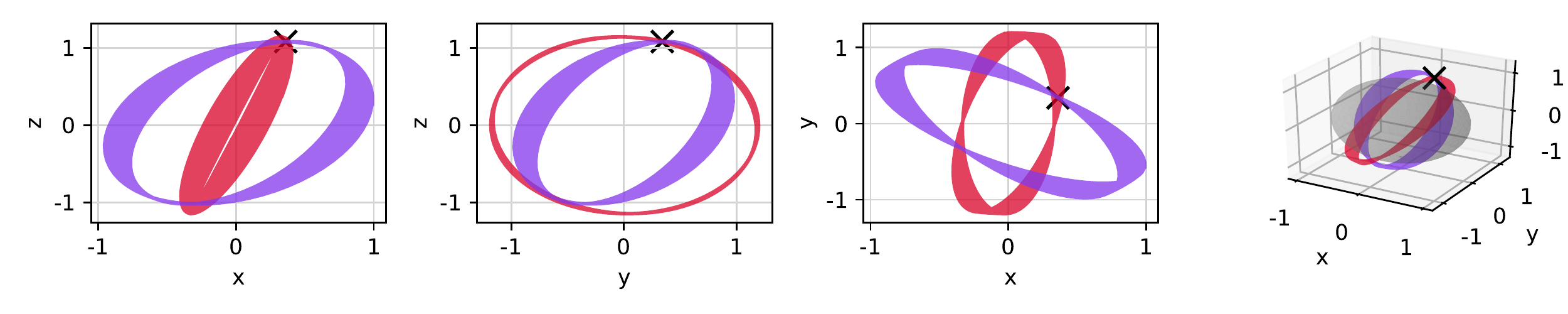}
\caption{Orbits of the target and chaser in the conjunction ground truth event that was used as a test case. ``X'' marks the conjunction.}
\label{fig:orbits}
\end{figure}

We run experiments on synthetic CDMs to evaluate our pipeline: first we sample a synthetic event from our model, generating a series of CDMs (for a single simulated conjunction) that we designate as observed data for subsequent inference; next, we perform posterior inference over model latents conditioned on the observed CDMs and compare resulting distributions with the known ground truth values of latents that produced the test observation. We use Gaussian likelihoods $\prod_{y \in \mathbf{y}} \mathcal{N}(\mu_y, \sigma_y)$, where $\mu_y$ is the value of each simulated observable and $\sigma_y$ is an empirically-tuned tolerance factor, taking $\mathbf{y}$ to include the TCA and observed object states transformed to Keplerian elements (eccentricity, inclination, semi-major axis). Figure \ref{fig:dists} shows a result using importance sampling with 70,000 samples, where we pick a selected subset of latent variables $\mathbf{x}$, namely the six initial orbital elements of target and chaser (i.e., mean motion, eccentricity, inclination, mean anomaly, argument of perigee and right ascension of ascending node). The presented results were conditioned on the first CDM of a test collision event with 16 CDMs (Figure~\ref{fig:orbits}). The ground truth values are shown with a dashed vertical line in Figure \ref{fig:dists}. As we observe, some distributions are of multimodal nature, suggesting that the estimated posterior can provide insight into multiple different explanations of the observed values. It is interesting to notice how the initial right ascension of the ascending node, argument of perigee, and mean anomaly, do not seem to play a relevant role in explaining the conjunction event. On the other hand, the mean motion, inclination, and eccentricity have posterior distributions that differ significantly from the priors, underlying their pivotal roles in explaining the observed conjunction event. The less relevant role of the first three orbital elements is to be expected since these orbits are almost circular (i.e., both chaser and target have very low eccentricity values). The probabilistic model has been implemented in Python and integrated with Pyprob,\footnote{\url{https://github.com/pyprob/pyprob} (September 2020)} a universal probabilistic programming framework \cite{baydin2019efficient}. We are currently investigating a range of likelihood formulations motivated by the physics of the problem and running experiments with Markov chain Monte Carlo \cite{brooks2011handbook} inference engines in PyProb that provide rich diagnostics tools to assess the convergence of the approximate posteriors. Moreover, due to the unavailability of public data, we are also working with space operators on evaluating this technique with real CDMs. This will enable us to further investigate and quantify the benefits of such methods on real-world applications.

\section{Conclusions}
In this preliminary work, we have introduced a novel probabilistic model to simulate conjunction events and the issuing of CDMs associated with each event. The model is calibrated with real data and the prior distributions have been selected to represent the current population of resident space objects in the LEO environment. Our work constitutes the first probabilistic model of synthetic CDMs for collision events by sampling from realistic prior distributions of resident space objects in LEO. 

This model can be leveraged for several purposes, on which we are currently working on: (1) Generating the first large dataset of synthetic CDMs calibrated with real data, contributing much needed, large-scale, public data to foster further work by the space and ML communities; (2) Investigating the joint probability space to understand which key variables and configurations lead to, and are useful to predict and prevent, conjunction events; (3) Creating a new generation of risk assessment tools using a Bayesian model-based framework, providing posterior distributions over event characteristics and future uncertainty evolution conditioned on real data observations that space operators have access to during an ongoing event.

\section*{Broader Impact}

Research that addresses the risk assessment and prevention of collisions in space is of significant short- and long-term benefit to humanity. A Kessler syndrome scenario where a chain reaction of collisions render low Earth orbit inaccessible would be a major environmental catastrophe that would significantly hinder space access for many generations. We believe that simulation-based inference and machine learning will be essential additions to the tool set for addressing these issues in a principled way. We do not foresee any negative societal impacts or harmful use of this research given the current circumstances.

\section*{Acknowledgments}

This work has been enabled by Frontier Development Lab (FDL) Europe, a public--private partnership between the European Space Agency (ESA), Trillium Technologies and the University of Oxford, and supported by Google Cloud. We would like to thank Dario Izzo and Moriba Jah for sharing their technical expertise and James Parr, Jodie Hughes, Leo Silverberg, Alessandro Donati for their support. AGB is supported by EPSRC/MURI grant EP/N019474/1 and by Lawrence Berkeley National Lab.

\pagebreak
\bibliographystyle{unsrt}
\bibliography{main.bib}

\end{document}